# CHINESE MEDICAL QUESTION ANSWER MATCHING BASED ON INTERACTIVE SENTENCE REPRESENTATION LEARNING

Xiongtao Cui and Jungang Han

College of Computer and Engineering, Xi'an University of Posts and Telecommunications, Xi'an, China

*ABSTRACT*

*Chinese medical question-answer matching is more challenging than the open-domain question-answer matching in English. Even though the deep learning method has performed well in improving the performance of question-answer matching, these methods only focus on the semantic information inside sentences, while ignoring the semantic association between questions and answers, thus resulting in performance deficits. In this paper, we design a series of interactive sentence representation learning models to tackle this problem. To better adapt to Chinese medical question-answer matching and take the advantages of different neural network structures, we propose the Crossed BERT network to extract the deep semantic information inside the sentence and the semantic association between question and answer, and then combine with the multi-scale CNNs network or BiGRU network to take the advantage of different structure of neural networks to learn more semantic features into the sentence representation. The experiments on the cMedQA V2.0 and cMedQA V1.0 dataset show that our model significantly outperforms all the existing state-of-the-art models of Chinese medical question answer matching.*

## 1. INTRODUCTION

In recent years, more and more patients seek answers through the online medical health community, which brings time convenience to patients and accumulates a large number of medical questions and answers data. The deep learning method can learn knowledge from the huge questions-answers dataset before automatically answer the question raised by patients[1], which not only shortens the waiting time of patients in queue, but also reduces the workload of doctors.

This paper focuses on the study of Chinese medical question-answer matching, which is a crucial step to automatically answer patient questions. For example, as shown in Table 1, for a patient's question, the question-answer matching is to select the most matched relevant answer from the candidate answer set. The Chinese question-answer matching in the field of professional medicine is more challenging than in the open-domain [2]. Due to the differences between medical and open-domain in thesaurus and word interpretation, existing word segmentation tools inevitably produce errors in the Chinese language processing of medical texts, which reduce the accuracy of question-answer matching. Zhang et al. [3] proposed a character-level embedding method to effectively solve the problem of word segmentation on medical text, and proposed a multi-scale interactive network framework in the later study to mine the semantic information and semantic association of medical questions and answers [4]. However, their model has limited performance in capturing semantic information and semantic association, which makes it difficult to proceed to practical application.

Table 1 An example of Chinese medical question-answer matching

| | |
|---|---|
| Question | 喉咙总有异物感感觉有痰一样咽不下去咳不出来，喉结左边一咳痰的时候也会跟着疼，但不会疼的很厉害，但是疼痛感明显，这是怎么回事啊医生？<br>There is always a foreign body feeling in the throat like phlegm can not be swallowed out cough, the left side of the throat knot when expectoration will follow the pain, but will not hurt very much, but the pain is obvious, what is the matter ah doctor? |
| Relevant answer | 这种情况考虑是属于慢性咽峡炎，可以配合医生进行相关调理治疗，比如清淡饮食，多喝水，必要时还可以进行雾化吸入以及口服适当的药物治疗，坚持治疗会有一定的效果，可以慢慢好转的。<br>This situation is considered to be chronic angina, you can cooperate with doctors for relevant conditioning treatment, such as light diet, drink more water, if necessary, you can also carry out atomization inhalation and oral appropriate drug treatment, adhere to the treatment will have a certain effect, can slowly improve. |
| Irrelevant answer | 通过你的描述，这种情况最好到医院化验一下血常规，看是细菌还是病毒引起的。<br>According to your description, it's better to go to the hospital for a blood test to see if it is caused by bacteria or viruses. |

In response to the above problems, we design a series of interactive sentence representation learning models. In these models, we propose an crossed BERT [5] network, which is a modification of the Siamese [6] structure using the BERT network. This makes the question and answer pay attention to each other's semantic information in the process of model learning, and sentence representation obtains more information features. Then, we add multi-scale CNNs [3] or bidirectional GRU [7] network into the model, to take the advantages of different neural network structures. Due to the strict professional requirements for answering medical questions in Chinese, we chose to conduct experiments on the cMedQA V2.0 and cMedQA V1.0 dataset. The experimental results show that our models significantly outperform the existing state-of-the-art models. The top-1 accuracy on development dataset and test dataset of cMedQA V2.0 dataset is improved by 9.2% and 10.1% respectively, and the top-1 accuracy on development dataset and test dataset of cMedQA V1.0 was improved by 10.2% and 9.8% respectively.

The other parts of this paper are organized as follows: Section 2 introduces the research work related to this paper; Section 3 proposes a series of models that we designed; Section 4 describes the data set cMedQA V2.0 and cMedQA V1.0 and analyses the experimental results; Section 5 summarizes our study.

## 2. RELATED WORK

We will briefly introduce the recent research works on the application of deep learning technology to question-answer matching in general fields and the professional medical field.

## 2.1. General Field

Early question answering methods such as logical rules [8] [9], information retrieval [10-12], and matching-based [13-15] only mined the shallow text information without extracting the deep semantic information of the text.

In recent years, more and more researchers have begun to focus on deep learning methods to mine deep text features of the text. Hu et al. [16] proposed a convolutional neural network model, which captures rich multi-level features within sentences to match two sentences. Qiu et al. [17] used a convolutional neural tensor network to model the interaction of two sentences through a tensor layer. Yin et al. [18] proposed an attention convolution neural network to model sentence pairs. Wang et al. [19] and Tan et al. [20] used an LSTM network to capture the order information of sequences while encoding sentences. Chen et al. [21] described a model based on position attention recurrent neural network to incorporate the word position of context into the attention representation. Wang et al. [22] added external attention information to hidden representation based on the recurrent neural network to obtain sentence representation containing attention. Tran et al. [23] presented a multi-hop attention mechanism, which uses multiple attention steps to learn the representation of the candidate answers.

The aforementioned work takes the advantages of neural network in extracting deep-level semantic features in the general field for question answering. However, answering questions in the specific medical field needs special study to void performance decline.

## 2.2. Medical Field

Compared with the general field, there is only a small amount of research work in Chines medical question answering. Perhaps the special Chinese language structure and the medical expertise complicated the problem.

Zhang et al [3] constructed the data set cMedQA V1.0 for Chinese medical question answering and proposed a multi-scale convolutional neural network model based on character embedding to extract text semantic information. They performed experiments on cMedQA V1.0, showed that character embedding and multis-cale convolutions were more advantageous than statistical rule methods.

Ye et al. [24] proposed a multi-layer composite convolutional neural network model, which stacks multiple convolutional neural networks together and extracts the characteristics of questions and answers from each layer, thus enriching the information of the final representation vector. They performed experiments on cMedQA V1.0 datasets and achieved the state-of-the-art performance at the time.

Zhang et al. [25] proposed a hybrid model of CNN and GRU neural networks for Chinese medical question-answer selection. Their model combines the advantages of different structures of two neural networks, thus achieving the most advanced performance on cMedQA V1.0 datasets.

Zhang et al. [4] proposed a method of incorporating attention mechanisms into multi-scale convolutional networks to focus on the interaction of questions and answers. And they constructed the cMedQA v2.0 data set, which is the optimization and update of cMedQA V1.0. The experimental results on the two data sets showed the advanced performance of their methods and that cMedQA v2.0 can better adapt to the complex neural network model.

Tian et al. [26] constructed a Chinese medical Q&A corpus called ChiMed and proposed a baseline model based on CNN and LSTM to validate this data set. He et al.[27] constructed a large-scale Chinese medicine question-answer dataset called webMedQA and proposed the convolution semantic clustering representation method to solve the question-answer matching problem.

The aforementioned methods have increasingly improved the performance in Chinese medical question-answer matching, but they are still limited in mining complex deep semantic information and the semantic association of question-answer pairs. Therefore, we aim to design more complex neural network models to overcome the limitation.

## 3. MODELS

The similarity between the sentence representations of questions and answers can measure their matching relationship, but it is only limited in the semantic information inside the sentence. Interactive sentence representation can focus on the connection between the questions and answers sentences to improve the accuracy of matching. Therefore, we designed four interactive sentence representation learning models with different architecture to extract more sentence features and capture connections of questions and answers. First, we construct a Siamese structured network model for question-answer matching using the BERT network, in which the question and the answer are represented as vectors of the same length for cosine similarity calculation. Then, we modified the neural network part of the previous model to a Crossed BERT network, which not only contains deep information features in the sentence representation, but also learns the semantic relationship between the question and the answer. Finally, we add a multi-scale CNNs network or bidirectional GRU network to the neural network part of the previous model, and their advantages in network structure can further extract more useful information features. In the following subsections, we will describe more technical details of these models.

### 3.1. Siamese BERT Model

The Siamese [6] BERT network model for question-answer matching is shown in Figure 1. The Siamese structure of this model includes two branches which share weights and parameters. The questions and answers will be represented as vectors of the same length for similarity calculation.

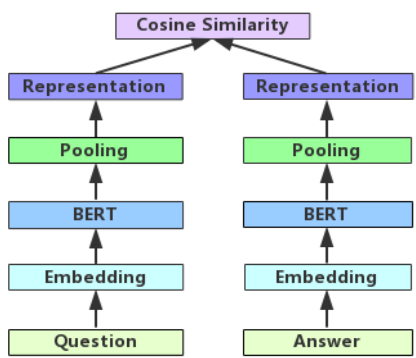

Figure 1. Siamese BERT network model.

The model uses the BERT network to learn sentence representation, as shown in Figure 2. We use position embedding, segment embedding, and token embedding of sentence sequences as inputs to the BERT network, where token is equivalent to Chinese characters. The character [CLS] is inserted into the sequence of sentences as the first token of a sentence and the character [SEP] as the last token of a sentence. The BERT network consists of multiple bidirectional transformer [29] encoder layers. In each layer, there is a multi-head self-attention sublayer, which pays attention to the connection between two words at any position, as shown in Figure 3. The output of the BERT network is the context encoding of the input sequence, which is denoted by following equation:

$$H = BERT(E_0, E_1, ..., E_n) \quad (1)$$

where $H = [T_0, T_1, ..., T_n]$, and $T_i$ are context representation of each token. They are input to the mean pooling layer to extract useful information, shown as:

$$P = Pool_{mean}(H) \quad (2)$$

where $P$ denotes the pooled output.

We use the pooling of BERT network output as a sentence representation. If the sentence representations of questions and answers are expressed as q and a respectively, then the cosine

similarity for calculating the association between q and a is shown as:

$$Sim(q, a) = \cos(q, a) = \frac{\|q \bullet a\|}{\|q\| \bullet \|a\|} \quad (3)$$

where ‖·‖ stands for vector length.

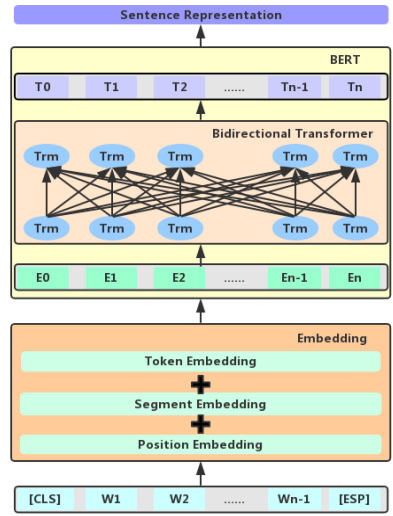

Figure 2. BERT for Sentence Representation.

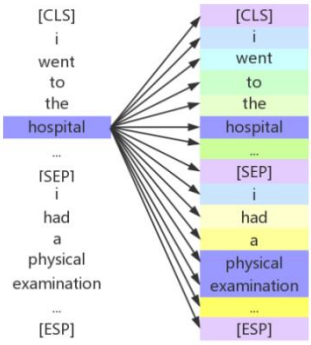

Figure 3. Self-attention sublayer.

## 3.2. Crossed BERT Siamese Model

The above model can extract deep-seated sentence representations of questions and answers separately, but the semantic correlation between them is ignored. Sergey et al. [28] proposed a 2-Channel network structure in the application of comparing the similarity of image patches, which inspired us to design the Crossed BERT Siamese network model, as shown in Figure 4.

In this model, where bidirectional transformer of two BERT networks interact with each other, as shown in Figure 5. Compared with the previous model, the neural network part was modified to an crossed BERT network. Therefore, the token output of question will pay attention to the Chinese characters in the answers and enrich the sentence representation of the questions. The token output calculation formula for the BERT of the questions is as follows:

$$T_{q_i} = Trm^{q_i}(E_{q_1}, E_{q_2}, \ldots, E_{q_n}, E_{a_1}, E_{a_2}, \ldots, E_{a_n}) \quad (4)$$

where $T_{q_i}$ is i-th output of the BERT network in the question, and $Trm$ is the bidirectional transformer. Likewise, the token output of the answer also pay attention to the semantic features in the question.

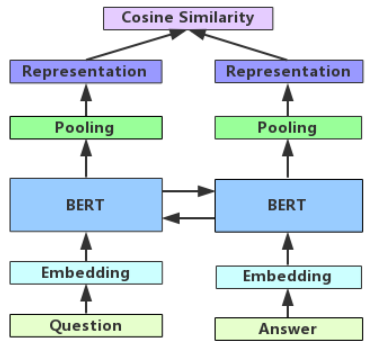

Figure 4. The Crossed BERT Siamese network model.

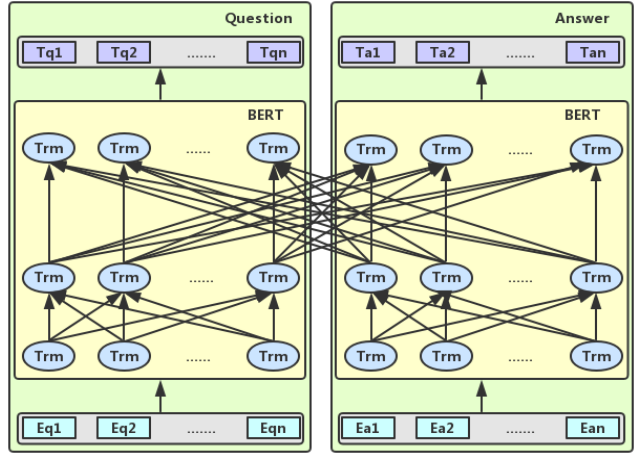

Figure 5. Crossed BERT network.

## 3.3. Crossed BERT Siamese Multi-Scale CNNs Model

Zhang et al [25] developed a hybrid model using CNN and GRU, combining the advantages of different neural network structures. Their method inspired us to design a hybrid model architecture of Crossed BERT Siamese network and Multi-Scale Convolutional Neural Networks (Multi-Scale CNNs), as shown in Figure 6.

In this model, the multi-scale CNNs network uses a series of convolution kernels of different sizes in convolution operations, each of which extracts n-gram features in sentences, as shown in Figure 7. Given a sequence $C=[t_0, t_1, ..., t_{l-k_i+1}]$ and convolution kernel size set $K = \{k_1, k_2, ..., k_s\}$, where the convolution output of the i-th convolution kernel $k_i$ is shown as：

$$O_j^{k_i} = f(W_j^{k_i} \circ [t_0, t_1, ..., t_{l-k_i+1}] + b^{k_i}) \tag{5}$$

where $O_j^{k_i} \in R^{l-k_i+1}$, $f(\cdot)$ is the activation function, $W_j^{k_i}$ are the matrix of weight parameters, vector $b^{k_i}$ is bias parameters, and $W \circ C$ is matrix multiplication. The number of convolution kernel is expressed as *N*, and the output of multi-scale CNNs network layer is $O^{k_i} = [O_0^{k_i}, O_1^{k_i}, ..., O_N^{k_i}]$. After that, we choose the max pooling to extract the useful feature information after convolution, the maximum value is more sensitive to the combined matrix features. Shown as:

$$p^{k_i} = [\max(O_0^{k_i}), \max(O_1^{k_i}), ..., \max(O_N^{k_i})] \tag{6}$$

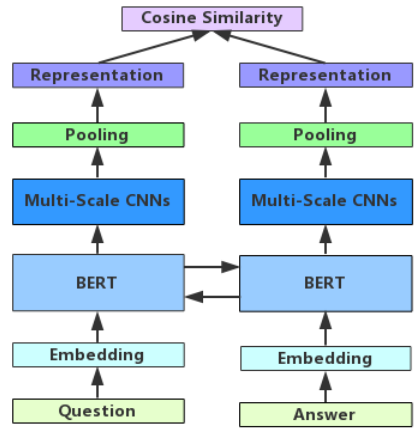

Figure 6. The Crossed BERT Siamese multi-scale CNNs.

Next, the convolution outputs of different scales are concatenated and expressed as $P = [p^{k_1}, p^{k_2}, ..., p^{k_S}]$.

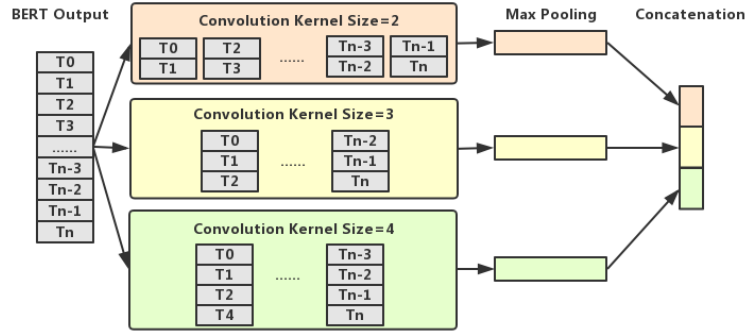

Figure 7. The multi-scale CNNs.

We represent the output of BERT network as $H=[T_0,T_1,...,T_n]$. Then, $H$ are input into the multi-scale CNNs network as shown in Figure 7. Finally, the output of multi-scale CNNs, P is used as sentence representation for cosine similarity calculation.

## 3.4. Crossed BERT Siamese BiGRU Model

CNN network is good at mining the static features of local position in sentences, but it is difficult to extract the order information of Chinese characters in sentences. The recurrent neural network (RNN) can capture sequence information in sentences. However, RNN may have problems with gradient disappearance and gradient explosion during model training [30]. The GRU network not only solves the problems of RNN, but also simplifies long-term short-term memory (LSTM) network and improves the model computing performance [7]. Therefore, we add a bidirectional GRU (BiGRU) network layer to the Crossed Siamese BERT model, as shown in Figure 8.

The hidden state of GRU network is shown in Figure 9. The hidden layer updates it state $h_t$ as shown below:

$$r_t = \sigma(W_r \bullet [h_{t-1}, x_t]) \tag{7}$$

$$z_t = \sigma(W_z \bullet [h_{t-1}, x_t]) \tag{8}$$

$$\tilde{h}_t = \tanh(W \bullet [r_t * h_{t-1}, x_t]) \tag{9}$$

$$h_t = (1 - z_t) * h_{t-1} + z_t * \tilde{h}_t \tag{10}$$

where $z_t$, $r_t$, σ, and $W$ are update gate, reset gate, sigmoid activation function and weight parameters respectively. The value of update gate ranges 0 to 1, which determines the memory of the previous hidden state in the current hidden state. The reset gate controls the amount of information entered into the current hidden state from the previous hidden state.

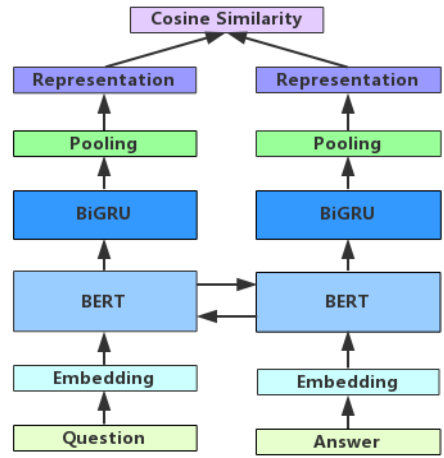

Figure 8. The Crossed BERT Siamese BiGRU.

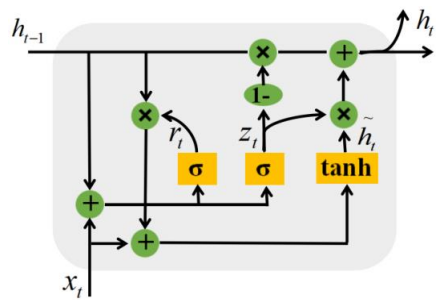

Figure 9. The hidden state of GRU network.

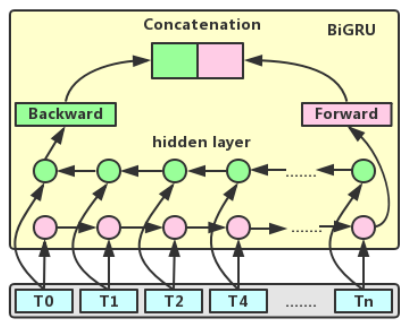

Figure 10. Network Structure of BiGRU network.

In this model, first inputs the question and answer to the BERT network to extract the features $H=[T_0,T_1,…,T_n]$. Then, $H$ are input into the BiGRU network layer of the two branches respectively. The network structure of BiGRU is shown in Figure 9. The forward output of the GRU network is $\vec{h}$ and the backward output is $\overleftarrow{h}$, then the BiGRU output is h = $\vec{h}$ ∥ $\overleftarrow{h}$. The output of BiGRU network layer can be shown as：

$$G = BiGRU(H) \tag{11}$$

Where $H \in R^{n \times 2h_d}$, and $h_d$ is hidden layer dimension. Then, we average pooling the output of the BiGRU network layer, shown as:

$$P = Pool_{mean}(G) \tag{12}$$

Finally, we use $P$ as sentence representation for cosine similarity calculation.

### 3.5. Objective Function

In this paper's model, we mark each tuple training data as $(q_i, a_i^+, a_i^-)$, where the relevant answer to question $q_i$ is $a_i^+$ and the irrelevant answer is $a_i^-$. The goal of model training is to maximize the cosine similarity between $q_i$ and $a_i^+$, and also minimize the cosine similarity between $q_i$ and $a_i^-$. We use margin loss function [31] as the training objective function of the model, which is defined as

$$L = \max\{0, M - Sim(q_i, a_i^+) + Sim(q_i, a_i^-)\} \tag{13}$$

where the margin value $M$ is a constant and represents the distance between $a_i^+$ and $a_i^-$. $Sim(\cdot)$ is the similarity of cosine. If the value of loss function is 0, then $M < |Sim(q_i, a_i^+) - Sim(q_i, a_i^-)|$. After that, we use fixing weight decay regularization in Adam (AdamW) [32] algorithm to update the training parameters of the model, which improves the generalization ability of the adaptive gradient algorithm. The algorithm will automatically reduce the learning rate along with the increasing of training time.

## 4. EXPERIMENTS

### 4.1. Dataset

We use the Chinese medical questions and answers dataset cMedQA V2.0 and cMedQA V1.0 to verify the effectiveness of our model in medical question answer matching task. The cMedQA V2.0 dataset was constructed by Zhang et al. [4] and was derived from an online Chinese medical health community(*http://www.xywy.com/*) which is provided by real users. The average number of characters for questions and answers in the cMedQA V2.0 dataset is 49 characters and 101 characters respectively, and the specific statistical results are shown in Table 2. The cMedQA V1.0 dataset is the initial version of cMedQA V2.0, which was collected by Zhang et al [3] and the detailed statistics are shown in Table 3.

### 4.2. Metrics

We used top-k accuracy (ACC@K) to evaluate the performance of our model, which is defined as

$$ACC@K = \frac{1}{N}\sum_{i=1}^{N} 1[a_i \in c_i^k] \tag{14}$$

where $c_i^k$ is the set of top-k answers with the highest similarity to the question $q_i$, which belongs to the candidate answer set. The expression $1[\bullet] \rightarrow \{0,1\}$ denote a mapping that if the value in square brackets is true then the mapped value is 1, otherwise 0.

Each question in the cMedQA V2.0 dataset or cMedQA V1.0 dataset has 100 candidate answers, and we used top-1(ACC@1) to evaluate the performance of our model. The random selection has an accuracy of only 1%, so this is a very stringent measurement.

Table 2. The statistics of cMedQA V2.0 dataset.

|  | Question | Answer | Average Characters Per Question | Average Characters Per Answer |
| --- | --- | --- | --- | --- |
| Train | 100,000 | 188,490 | 48 | 101 |
| Development | 4,000 | 7,527 | 49 | 101 |
| Test | 4,000 | 7,552 | 49 | 100 |
| Total | 108,000 | 203,569 | 49 | 101 |

Table 3. The statistics of cMedQA V1.0 dataset.

|  | Question | Answer | Average Characters Per Question | Average Characters Per Answer |
| --- | --- | --- | --- | --- |
| Train | 50,000 | 94,134 | 120 | 212 |
| Development | 2,000 | 3,774 | 117 | 216 |
| Test | 2,000 | 3,835 | 119 | 211 |
| Total | 54,000 | 101,743 | 119 | 212 |

### 4.3. Baselines

To evaluate the performance of our model, we use the baseline models of the related studies as follows：

- **Single-CNN**: The model of Siamese structure, there is only one size convolution kernel to handle the question answer matching task.
- **Multi-Scale CNNs:** The model in which different scales of convolution are used in Siamese network to capture deep semantic information of questions and answers [3].
- **Multi-Level Composite CNNs:** The model proposed by Ye et al. [24] to extract intermediate features from each layer of convolution of Multi-layer CNNs, not just the superposition of multi-layer networks.
- **BiGRU**: The model in which the Siamese structure of the BiGRU network was used to capture the deep semantics and dependencies of question-and-answer pairs.
- **BiGRU Multi-Scale CNNs**: The model proposed by Zhang et al. [25] with multiple network hybrid structures and achieved state-of-the-art performance on cMedQA V1.0 datasets. The model takes the output of BiGRU as the input of multi-scale CNNs. It can capture not only local location invariant features but also sequence and dependent information.
- **BiGRU Shortcuts Multi-Scale CNNs Interactive**: The multi-scale interaction model proposed by Zhang et al [4] designed with shortcuts connection. The output of BiGRU and the previous embedded vectors are sent to the multi-scale CNNs, and then the attention interaction matrix is generated as the weight of the pooled output vector.

## 4.4. Experimental parameters

Our model is built using Pytorch framework. We conducted the experiments using DGX-1 deep learning server from Nvidia Corporation. In order to reduce the training time, we use 80% of the training dataset for model training.

All of our models used the pre-trained BERT model, which is the Chinese version of Google's BERT-Base model [5]. We pre-trained BERT-Base model using the Chinese Medical Corpus. The BERT-Base model has 12 transformer layers, 768 hidden states and 12 heads with self-attention, totaling 110M parameters. We process the input of the model, which treats the length of the sequence of questions and answers into a fixed length of 150 Chinese characters. If the length is less than 150 Chinese characters, padding the remaining positions with zero. If there are more than 150 Chinese characters in the sentence, it will be truncated. The feature maps for each convolution scale of Multi-Scale CNNs are 500. The output of BiGRU in each direction is 200 in dimensions. The margin value $M$ of the loss function is 0.1. The initial learning rate is 2e-5.

## 4.5. Results

The experimental results of our model on the cMedQA V2.0 and cMedQA V1.0 dataset are shown in row -12 of Table 4. The Crossed BERT Siamese multi-scale CNNs use the convolution kernel with size 2 and 3. Dev (%) is the top-1 accuracy of development set. Test (%) is top-1 accuracy (ACC@1) of the validation set.

Models in rows 1 to 8 of table 4 list the performance of baseline models. Among the first three single network models, Multi-Scale CNNs achieves the highest computational score for model evaluation, which can capture semantic feature information with different size of granularity. The baseline models below the three models are multi-layer network structures. Compared with the previous three single network models, the evaluation scores were slightly improved. The combination of BiGRU and CNN shows that the multi-network layer model is feasible. Multi-Level Composite CNNs extract feature information from each convolution layer, enriching the final vector representation feature. The BiGRU network shortcuts Multi-Scale CNNs interactive model not only mix the advantages and disadvantages of the previous model, but also focuses on the interaction of questions and answers in the training process, so it improves the performance of the model.

The models in row 9 to 12 are four Crossed sentence representation network models proposed in this paper. Compared with Siamese BERT model, the performance of the Crossed BERT Siamese model is significantly improved. This indicates that the relationship between sentences should be paid attention in the question answer matching task. Comparing rows 10 and 11, the top-1 accuracy of the Crossed BERT BiGRU model is higher. This demonstrates that BiGRU has an advantage in compensating for the deficiencies of BERT network.

Table 4. Top-1 accuracy (ACC@1) results of model.

| Model | cMedQA V2.0 | | cMedQA V1.0 | |
|---|---|---|---|---|
| | Dev(%) | Test(%) | Dev(%) | Test(%) |
| CNN | 67.6 | 67.8 | 64.0 | 64.5 |
| BiGRU | 68.9 | 68.7 | 64.9 | 66.7 |
| Multi-Scale CNNs[3] | 70.0 | 70.9 | 65.4 | 64.8 |
| BiGRU-CNN | 69.5 | 70.0 | - | - |
| CNN-BiGRU | 67.9 | 67.7 | - | - |
| BiGRU Multi-Scale CNNs[25] | - | - | 68.4 | 68.4 |
| Multi-Level Composite CNNs[24] | 70.4 | 70.1 | 65.6 | 66.2 |
| BiGRU Shortcuts Multi-Scale CNNs Interactive[4] | 72.1 | 72.1 | 66.1 | 67.1 |
| Siamese BERT | 78.3 | 78.6 | 75.1 | 75.2 |
| Crossed BERT Siamese | 80.5 | 80.4 | 77.3 | 77.9 |
| Crossed BERT Siamese Multi-Scale CNNs | 80.2 | 80.7 | 77.5 | 77.6 |
| Crossed BERT Siamese BiGRU | **81.3** | **82.2** | **78.6** | **78.2** |

The above experimental results show that our series of models have better performance than all the baseline models, especially the Crossed BERT Siamese BiGRU model. An important reason for the improved performance of the Chinese medical question-answer matching model is our proposed Crossed BERT network. This also illustrates that question-answer matching requires not only abundant sentence features, but also relevant information between question-answer pairs.

### 4.6. Discussion

#### 4.6.1. Pre-training

In this paper, the pre-training model we used is a Chinese version of the BERT-Base model provided by Google [5]. The BERT-Base model uses Chinese corpus training in the general field. In medical field, text structure is more complex and word combination is diversified. Therefore, we collected a large number of medical texts to construct a medical corpus and used it to pre-train the BERT-Base model.

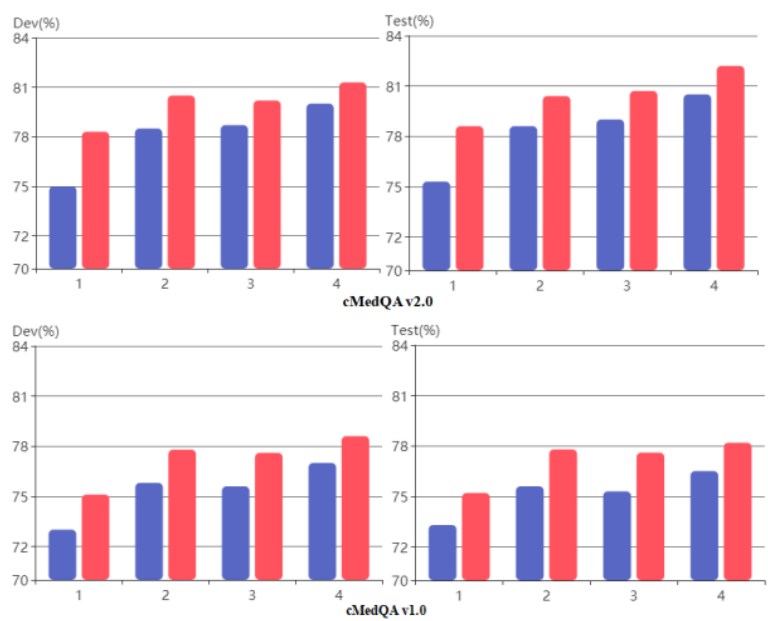

Figure 11. Results of pre-training experiment.

On the cMedQA V2.0 and cMedQA V1.0 dataset, we conducted a comparative experiment on the without pre-training BERT-Base model and Chinese Medical pre-training BERT model, as shown in Figure 11. The blue histogram represents the without pre-training BERT-Base model. The red histogram represents Chinese Medical pre-training BERT model. Dev (%) is the top-1 accuracy of development set. Test (%) is top-1 accuracy (ACC@1) of the validation set. The abscissa 1 represents the Siamese BERT model, the abscissa 2 represents the Crossed BERT

Siamese model, the abscissa 3 represents the Crossed BERT Multi-Scale CNNs model, and the abscissa 4 represents the Crossed BERT Siamese BiGRU model.

The experimental results depicted in figure 11 show that Chinese Medical pre-training BERT model performs better than the without pre-training BERT based model. Therefore, to use the pre-training model for sentence representation learning for Chinese medical texts, it is necessary to take into account the differences between the medical field and the general field and adapting to current language tasks.

### 4.6.2. Sentence representation of BERT network

In order to adapt the BERT network to sentence representation learning, we have carried out experiments and discussions on the output of the BERT network in the Siamese BERT model. We choose three methods of BERT network output commonly used for language tasks:

1) First Token: Take the first token of the last layer of BERT network as output directly;
2) Mean Token: The average pooling of all tokens at the last layer of the BERT network is used as output;
3) Mean Useful Token: The padding position of the last layer of the BERT network is covered with zero, and then the useful tokens average pooling is used as the output.

The experimental results on cMedQA V2.0 dataset and cMedQA V1.0 dataset are shown in Table 6. The first column represents the output category of the BERT network in the Siamese Bert model. The Following column, Dev(%) is the top-1 accuracy of development set. Test(%) is top-1 accuracy(ACC@1) of the validation set.

Table 6. Results of BERT network output.

| BERT Output | cMedQA V2.0 | | cMedQA V1.0 | |
| --- | --- | --- | --- | --- |
| | Dev(%) | Test(%) | Dev(%) | Test(%) |
| First Token | 73.45 | 73.95 | 70.39 | 70.95 |
| Mean Token | 77.30 | 77.78 | 74.05 | 74.14 |
| Mean Useful Token | **78.28** | **78.58** | **75.15** | **75.25** |

The above experimental results show that mean useful token of BERT performs better on cMedQA V2.0 dataset and cMedQA V1.0 dataset. Therefore, the application of BERT network in sentence representation learning needs more abundant and useful features.

### 4.6.3. Error Analysis

Our proposed model is significantly superior to previous state-of-the-art models and achieves better performance on cMedQA V1.0 and cMedQA V2.0 datasets. However, in the practical application of automatic answering to medical questions, the guaranteed accuracy is still not achieved. Three reasons may lead to this problem. The first reason is the uneven distribution of experimental datasets and the more complex sentence semantics in the medical field. The second reason is the error generated while collecting data sets manually. The third reason is that a question may correspond to multiple answers, and our model only matches a single answer, which affects the final performance of our model.

## 5. CONCLUSIONS

In this paper, we propose a series of sentence representation learning models for Chinese medical question-answer matching. They can not only capture deeper semantic information from question sentences and answer sentences, but also integrate the association information into the final representation vector. The experimental results on the cMedQA V2.0 and cMedQA V1.0 dataset show that our model achieves better performance than that of all the baseline models.

Zhang et al [33] proposed a new graph neural network graph BERT, which improved the performance and computational efficiency of the traditional graph neural network. Li et al [34] designed a graph matching network which is used to calculate the similarity between two graph structure data. Inspired by their research, as the future work, we will try to transform unstructured text data into graph structure similar to a knowledge map. The data of graph structure is easier for a machine to understand, so as to improve the accuracy of our model in answering questions. And we will collect more medical question-and-answer data sets to improve the performance of the model.

Computer Science & Information Technology (CS & IT)    109

## AUTHORS

**Xiongtao Cui**

He received his bachelor's degree in network engineering from Xi'an University of Posts and telecommunications in 2018, and is currently pursuing a master's degree in big data processing and high performance computing. His research interests include machine learning, deep learning and natural language processing.

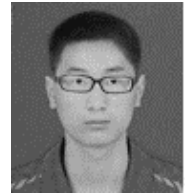

**Jungang Han**

He is a professor at Xi'an University of Posts and Telecommunications. He is the author of two books, and more than100 articles in the field of computer science. His current research interests include artificial intelligence, deep learning for medical image processing.

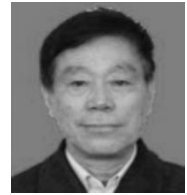